\documentclass[letterpaper]{article} 
\usepackage{aaai23}  
\usepackage{times}  
\usepackage{helvet}  
\usepackage{courier}  
\usepackage[hyphens]{url}  
\usepackage{graphicx} 
\usepackage{natbib}  
\usepackage{caption} 
\usepackage{amsfonts}
\usepackage{xcolor}         
\usepackage{booktabs}

\title{A Systematic Study on Quantifying Bias\\ in GAN-Augmented Data}
\author {
    Denis Liu\\
    Advised By: Alberto Olmo,
    Subbarao Kambhampati
}
\affiliations {
    School of Computing and AI, Arizona State University, Tempe\\
    dfliu@asu.edu, aolmoher@asu.edu, rao@asu.edu
}

\begin{document}

\maketitle

\begin{abstract}
Generative adversarial networks (GANs) have recently become a popular data augmentation technique used by machine learning practitioners. However, they have been shown to suffer from the so-called mode collapse failure mode, which makes them vulnerable to exacerbating biases on already skewed datasets, resulting in the generated data distribution being less diverse than the training distribution. To this end, we address the problem of quantifying the extent to which mode collapse occurs. This study is a systematic effort focused on the evaluation of state-of-the-art metrics that can potentially quantify biases in GAN-augmented data. We show that, while several such methods are available, there is no single metric that quantifies bias exacerbation reliably over the span of different image domains.
\end{abstract}

\section{Introduction}
Generative adversarial networks (GANs), introduced in ~\citet{GoodfellowPMXWO20}, are a type of neural networks that aim to replicate the distribution of the datasets they are trained on through generating new samples. This is celebrated by practitioners as a reliable and extensively used tool for data augmentation, especially in domains where data is scarce, as they have shown the seeming ability to learn the underlying data distribution with few examples, however complex.

However, due to an inherent failure mode called mode collapse~\cite{Goodfellow17}, when the training distribution is skewed towards some dimension, GANs have been proven to not only perpetuate but also magnify
the existing biases in the data. This makes the generated data distribution less diverse than the original, which could pose ethical implications in human-data datasets where biases pertaining to sensitive attributes such as race or gender can be affected~\cite{jain2021imperfect}.
This study aims to quantify bias by systematically evaluating various state-of-the-art metrics that quantify the extent to which mode collapse occurs in GAN-augmented datasets, with the motivation being that quantifying mode collapse could help elude adverse societal consequences that stem from the use of biased data.

\begin{figure*}[t]
\centering
\includegraphics[width=0.9\columnwidth]{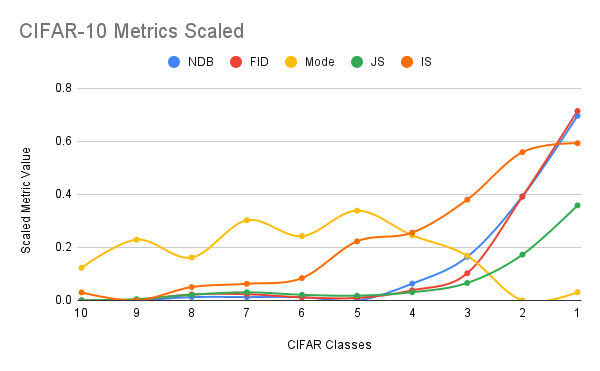} 
\includegraphics[width=0.9\columnwidth]{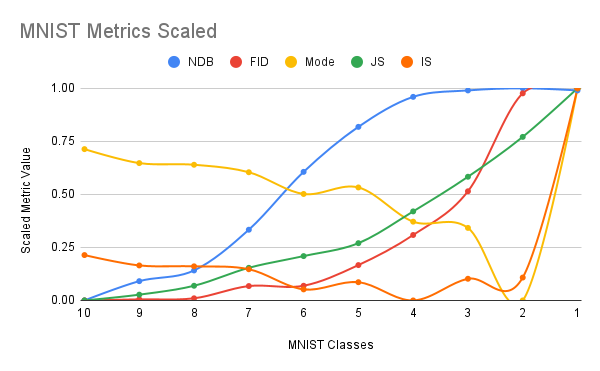} 
\caption{Scores obtained by the metrics over the artificially mode-collapsed datasets of CIFAR-10 (left) and MNIST (right). X-axis represents the number of classes left in the test set. Values are scaled between 0 (more diverse) and 1 (less diverse).}
\label{fig1}
\end{figure*}

\section{Methodology}
Five metrics were evaluated on the MNIST~\cite{lecun_bottou_bengio_haffner_1998} and CIFAR-10~\cite{Krizhevsky2009LearningML} datasets.
For each dataset, a set of ten artificially mode collapsed/skewed subsets were created by sampling from the original such that the $i$th dataset created (for $0\leq i\leq 9$) has classes $0, 1,\ldots i$. In addition to evaluating the five metrics on MNIST and CIFAR-10, each metric was also evaluated by testing them on GAN-generated datasets trained on MNIST and CIFAR-10 using DCGAN~\cite{RadfordMC15}, which was selected because of its popularity and tendency to exacerbate biases.

Our evaluation uses the following set of well-established metrics that measure the difference in diversity between two distributions.



\paragraph{Number of Statistically-Different Bins (NDB)}
    The NDB~\cite{10.5555/3327345.3327486} method uses K-means clustering to place samples into bins, after which a series of two-sample tests are performed to find the number of statistically-different bins. 
    As an alternative, the \textbf{Jensen-Shannon score} is used to find the divergence between the reference bins distribution and that of the tested model when the number of samples is high~\cite{10.5555/3327345.3327486}.
\paragraph{Inception Score (IS)} 
    The IS~\cite{NIPS2016_8a3363ab} obtains the conditional label distribution $p(y\mid x)$ for every image and calculates the entropy of the conditional label distribution by finding the KL divergence with the generated samples' label distribution denoted as $p(y)$ by finding
    \[\exp\left(\mathbb{E}_xD_{KL}(p(y\mid x)\parallel p(y)\right).\]
    
\subsubsection{MODE Score}
    The MODE score~\cite{Che2017ModeRG} alters the IS by incorporating the label distribution of the training samples, denoted as $p^*(y)$. Calculating it is as follows:
    \[\exp(\mathbb{E}_xD_{KL}(p(y\mid x)\parallel p^*(y)) - D_{KL}(p(y)\parallel p^*(y))).\]
    
\subsubsection{Fréchet Inception Distance (FID)}
    The FID~\cite{10.5555/3295222.3295408} compares the distribution of generated images to the distribution of real images. After finding the Guassian with mean $(m, C)$ and $(m_w, c_w)$ from the original and the generated distributions respectively, the FID is equivalent to
    \[\left\Vert m - m_w\right\Vert^2 + Tr\left(C + C_w - 2(CC_w)^{1/2}\right).\]

\section{Experiments and Results}

\begin{table}[]
\centering
\begin{tabular}{cccccc}
\toprule
\multicolumn{1}{l}{} & NDB  & FID  & MODE & JS   & IS   \\ \midrule
CIFAR-10                & 1.0    & 1.0    & 1.0    & 1.0    & 1.0    \\ \midrule
MNIST                & 0.89 & 0.22 & 0.50 & 0.74 & 0.02 \\ \bottomrule
\end{tabular}
\caption{Scaled results of the metrics obtained from the DCGAN-generated distributions when trained on CIFAR-10 and MNIST.}
\label{table1}

\end{table}


Experimental results are presented in terms of the five measures and two sets of data: (1) artificially mode-collapsed datasets and (2) DCGAN-generated images.

\paragraph{Artificially Mode Collapsed Datasets}
We artificially skew the distribution by removing one class at a time for both datasets and show the metrics' outputs in Figure~\ref{fig1}. Values are scaled between 0 and 1 and are inversely proportional to diversity. 

The NDB, JS, and FID scores performed similarly well for both tested datasets; they all increase monotonically according to the amount of bias. However, these fail to detect any variation in diversity for CIFAR-10 until at least \emph{six} classes are removed. The MODE score, on the contrary, does not show the extent of mode collapse on either dataset, showing inconsistent results for both dataset evaluations. 
The IS fares well in the CIFAR-10 tests as it tends to give worse scores to less diverse distributions. On the other hand, it fails to capture any lack of diversity in the MNIST experiments until the dataset consists of only one class.

\paragraph{DCGAN-Generated Images}
For each dataset, we train DCGAN, generate a set of 5000 samples, and evaluate them with each metric. The results, shown in Table \ref{table1}, depict low diversity scores from each metric for CIFAR-10. Upon inspection of the generated CIFAR-10 images, these results are sensible because only 2 classes were generated.
Contrastively, MNIST results are varying -- this is likely due to the noisy artifacts generated by DCGAN, affecting the performance of the measures. 
Potential future work is to examine the MNIST DCGAN dataset manually to estimate mode collapse and corroborate the effectiveness of the measures.

\section{Conclusions and Future Work}
Overall, the results show that measuring mode collapse remains an open challenge. Future work will involve two key items revolving around (1) the addition of other related measures and (2) the experimentation with higher quality benchmark datasets to prevent noisy outputs. Specifically, as the state-of-the-art is improved, more measures can be added to this proposed framework of evaluation. Further, adding labeled GAN-generated datasets already used in existing literature -- especially those that pose societal consequences such as ones involving face-imagery or human data -- would serve as benchmarks that more accurately reflect bias in society.

\bibliography{aaai23.bib}
\end{document}